%% file: paper.tex
\documentclass[acmsmall,screen,nonacm]{acmart}

\usepackage[capitalise]{cleveref}
\usepackage{siunitx}
\DeclareSIUnit\pixel{pixels}
\usepackage{acronym}
\input{acronyms.tex}
\usepackage{subcaption}

\usepackage{xcolor}
\definecolor{color1}{HTML}{057D9F} 
\definecolor{color2}{HTML}{FF8100} 
\definecolor{color3}{HTML}{E1004C} 
\definecolor{color4}{HTML}{70E500} 
\definecolor{color5}{HTML}{FFDB00} 
\definecolor{color6}{HTML}{CC00CC} 
\definecolor{color7}{HTML}{00CCCC} 

\usepackage{xspace}
\newcommand{\ie}{i.\,e.\xspace}

\newcommand{\eg}{e.\,g.\xspace}

\newcommand*\dlt[1]{\ensuremath{\mathrm{d}#1}}

\AtBeginDocument{%
  }

\begin{document}

\title{Vocalics in Human-Drone Interaction}

\author{Marc Lieser}
\email{marc.lieser@hs-rm.de}
\orcid{0000-0002-0379-7874}
\author{Ulrich Schwanecke}
\email{ulrich.schwanecke@hs-rm.de}
\orcid{0000-0002-0093-3922}
\affiliation{%
  \institution{RheinMain University of Applied Sciences}
  \streetaddress{Kurt-Schumacher-Ring 18}
  \city{Wiesbaden}
  \country{Germany}
  \postcode{65197}
}

\renewcommand{\shortauthors}{Lieser, Schwanecke}

\begin{abstract}
  \input{0_abstract}
\end{abstract}

\ccsdesc[500]{Human-centered computing~User studies}
\ccsdesc[500]{Human-centered computing~Auditory feedback}
\ccsdesc[500]{Computer systems organization~Robotics}
\ccsdesc[300]{Hardware~Sound-based input / output}

\keywords{human-robot interaction, human-drone interaction, robot sound, sonic interaction, non-lexical sounds, vocalics, social robots, drone communication}


\maketitle

\input{1_introduction}
\input{2_gestures}
\input{3_userstudy}
\input{4_discussion}
\input{5_conclusion}


\bibliographystyle{ACM-Reference-Format}
\bibliography{library}


\end{document}

%% file: acronyms.tex
\newacro{cvmr}[CVMR]{computer vision and mixed reality}
\newacro{hsrm}[HSRM]{high-speed and robust monocular}
\newacro{pzai}[PZAI]{Doctoral Center Applied Informatics}
\newacro{icarus}[ICARUS]{infrastructure for compact aerial robots under supervision}
\newacro{grasp}[GRASP]{general robotics, automation, sensing \& perception}
\newacro{fma}[FMA]{flying machine arena}
\newacro{ar}[AR]{augmented reality}
\newacro{bnf}[BNF]{bind and fly}
\newacro{rtf}[RTF]{ready to fly}
\newacro{dsmx}[DSMX]{digital spectrum modulation}
\newacro{ekf}[EKF]{extended Kalman filter}
\newacro{gps}[GPS]{global positioning system}
\newacro{ism}[ISM]{industrial, scientific and medical}
\newacro{mav}[MAV]{micro aerial vehicle}
\newacro{ppm}[PPM]{pulse-position modulation}
\newacro{sar}[SAR]{search and rescue}
\newacro{slam}[SLAM]{simultaneous localization and mapping}
\newacro{stem}[STEM]{science, technology, engineering, and mathematics}
\newacro{uart}[UART]{universal asynchronous receiver transmitter}
\newacro{uav}[UAV]{uncrewed aerial vehicle}
\newacro{ugv}[UGV]{uncrewed ground vehicle}
\newacro{rmse}[RMSE]{root-mean-square error}
\newacro{mse}[MSE]{mean squared error}
\newacro{ros}[ROS]{robot operating system}
\newacro{hri}[HRI]{human-robot interaction}
\newacro{hdi}[HDI]{human-drone interaction}
\newacro{mpc}[MPC]{model-predictive control}
\newacro{nmpc}[NMPC]{nonlinear model-predictive control}
\newacro{qp}[QP]{quadratic programming}
\newacro{sqp}[SQP]{sequential quadratic program}
\newacro{lq}[LQ]{linear quadratic}
\newacro{rk}[RK]{Runge-Kutta}
\newacro{tom}[ToM]{theory of mind}
\newacro{imu}[IMU]{inertial measurement unit}
\newacro{nui}[NUI]{natural user interface}
\newacro{gui}[GUI]{graphical user interface}
\newacro{dof}[DOF]{degrees of freedom}
\newacro{iss}[ISS]{international space station}
\newacro{sam}[SAM]{self-assessment manikin}
\newacro{vr}[VR]{virtual reality}
\newacro{spa}[SPA]{sparse bundle adjustment}
\newacro{vtol}[VTOL]{vertical take-off and landing}
\newacro{pcb}[PCB]{printed circuit board}
\newacro{esc}[ESC]{electronic speed controller}
\newacro{fpv}[FPV]{first-person view}
\newacro{faa}[FAA]{federal aviation administration}
\newacro{auw}[AUW]{all-up weight}
\newacro{fft}[FFT]{fast Fourier transform}
\newacro{stft}[STFT]{short-time Fourier transform}
\newacro{iae}[IAE]{integral absolute error}
\newacro{bfgs}[BFGS]{Broyden, Fletcher, Goldfarb, and Shanno}
\newacro{ode}[ODE]{ordinary differential equation}
\newacro{svd}[SVD]{singular value decomposition}
\newacro{lerp}[Lerp]{linear interpolation}
\newacro{slerp}[Slerp]{spherical linear interpolation}
\newacro{ewma}[EWMA]{exponentially weighted moving average}
\newacro{pid}[PID]{proportional-integral-derivative}
\newacro{api}[API]{application programming interface}
\newacro{sdk}[SDK]{software development kit}
\newacro{ir}[IR]{infrared}
\newacro{udp}[UDP]{user datagram protocol}
\newacro{pnp}[PnP]{perspective-n-point}
\newacro{ac}[AC]{alternating current}
\newacro{dc}[DC]{direct current}
\newacro{rc}[RC]{radio control}
\newacro{lipo}[LiPo]{lithium polymer}
\newacro{fov}[FOV]{field of view}
\newacro{tof}[ToF]{time-of-flight}
\newacro{rtt}[RTT]{round-trip time}
\newacro{emi}[EMI]{electromagnetic interference}
\newacro{bpm}[BPM]{beats per minute}
\newacro{sfu}[SFU]{semantic-free utterances}
\newacro{mfcc}[MFCC]{mel-frequency cepstral coefficient}
\newacro{anc}[ANC]{active noise cancellation}
\newacro{lti}[LTI]{linear time-invariant}

%% file: 0_abstract.tex
%
As the presence of flying robots continues to grow in both commercial and private sectors, it necessitates an understanding of appropriate methods for nonverbal interaction with humans.
While visual cues, such as gestures incorporated into trajectories, are more apparent and thoroughly researched, acoustic cues have remained unexplored, despite their potential to enhance human-drone interaction.
Given that additional audiovisual and sensory equipment is not always desired or practicable, and flight noise often masks potential acoustic communication in rotary-wing drones, such as through a loudspeaker, the rotors themselves offer potential for nonverbal communication.
In this paper, quadrotor trajectories are augmented by acoustic information that does not visually affect the flight, but adds audible information that significantly facilitates distinctiveness.
A user study ($N$=192) demonstrates that sonically augmenting the trajectories of two aerial gestures makes them more easily distinguishable.
This enhancement contributes to human-drone interaction through onboard means, particularly in situations where the human cannot see or look at the drone.

%% file: 1_introduction.tex
\section{Introduction}\label{sec:introduction}

As robots become increasingly integrated into our daily lives, it is crucial to ensure their effective and safe interaction between robots and humans.
Failure to meet human expectations in their responses can quickly lead to frustration.
Poorly implemented robot responses can negatively impact the acceptance of robots, turning otherwise positive encounters into unpleasant experiences.

The field of \ac{hdi} is dedicated to enhancing the relationship and collaboration between humans and flying robots in various interaction aspects, ultimately improving human-robot encounters in everyday environments.
It is essential for robots to comprehend human commands, while humans should intuitively understand a robot's intent.
The responsibility lies with robot designers and \ac{hri} researchers to craft social robots and provide appropriate interaction methods.
Established methods to indicate, \eg, a robot's status through visual cues like displays, are often applied to recent platforms.

There are, however, robots that are constrained in their appearances and where the adoption of such methods is more challenging.
Flying robots face significant limitations in payload capacity, often needing to prioritize equipment for flight-related functions or size reduction.
But despite their mechanical design, they can be retrofitted with features that enhance human acceptance and interaction experience, such as exploiting possible channels of nonverbal communication.

\subsection{Nonverbal Human-Drone Communication} 

The field of nonverbal \ac{hdi} is still emerging.
Due to their aerial nature, established concepts from, \eg, ground-based robots, are not directly transferable to flying robots.
Kinesics, proxemics, and haptics stand out as the most prominent research topics in the field.
Drone \emph{kinesics} researches how emotions or intent can be conveyed through the way the robots move~\cite{Cauchard2016,Karjalainen2017,Deng2018,lieser2021eave,bevins2021} which also includes drone gestures, \eg, to be utilized for pedestrian guidance~\cite{Colley2017}.
The field of drone \emph{proxemics} examines the extent to which people are comfortable with the proximity of drones and how close they tolerate them to their bodies~\cite{duncan2013,Yeh2017,acharya2017,Cauchard2019,lieser2021usertest,kunde2023}.
Drone \emph{haptics} is an area of research that studies direct interaction with drones, \eg, direct control with physical buttons attached to the drone~\cite{Rajappa2017}, virtual buttons~\cite{lieser2021metrodrone} utilizing the onboard \ac{imu}, or even the landing of drones on the human body~\cite{Auda2021}.

Research in the area of nonverbal, sound-based communication has increased over the past few years~\cite{Zhang2023,Pelikan2021}.
\emph{Vocalics}, the study of paralanguage, refers to the nonphonemic properties of speech, such as volume, pace, pitch, rate, rhythm, and tone.
Paralanguage offers great potential to enhance interaction as it conveys both meaning and emotion through vocal characteristics.
\Acp{sfu} in animation movies allow robots to communicate affect and intent~\cite{Yilmazyildiz2016} and inspired researchers to enhance robot communication by artificial movement sound~\cite{robinson2021}.
It also motivated efforts to generalize sound design recommendations~\cite{robinson2022} based on interviews with the designers of robotic consumer products like the Anki Vector, which uses non-lexical audio for communication~\cite{pelikan2020}.

\subsection{Motivation} 

Paralanguage has not yet found widespread adoption in the field of \ac{hdi}.
One possible reason for this is the rotor sound already emitted as a consequence of the drone's operation, referred to as consequential sound.
Being perceived as noise by humans, it negatively affects robot interaction experience~\cite{Cauchard2015,Jones2016,Colley2017,Chang2017,Knierim2018}.
For this reason, among others, the reduction of acoustic signature is the subject of current research, including the development of \ac{anc} solutions~\cite{Narine2020} and the exploration of new blade geometries such as the toroidal rotor~\cite{toroidal}.
However, as of now, neither of these technologies has reached the consumer stage.

To incorporate potential forms of acoustic communication, they have to compete against the drone's consequential sound.
Adding natural sounds, such as the twittering of birds or rain, through a loudspeaker attached to a drone increased interaction pleasantness but also the perceived loudness~\cite{Wang2023}.
There are, however, groups that can benefit from rotor noise, such as visually impaired people who can be aurally navigated using a quadrotor's rotor sound~\cite{Soto2015,Zayer2016,Soto2017,Soto2018}.
For this reason, shaping the consequential sound of quadrotors should be exploited and evaluated for acoustic communication.
Utilizing the noise sources that are already present as a consequence of the operation of rotary-wing drones is a minimalist approach that avoids the additional payload that even a small loudspeaker entails.
By supplementing gestures with vocalic cues, the drone does not always have to be kept in sight during interaction.
There are also scenarios in which the drone is not even intended to be in the human's view, \eg, when filming a jogger from behind.

This paper proposes a vocalic approach by augmenting a quadrotor's consequential sound with a higher-frequency oscillation while maintaining the visually perceived flight characteristics.
This method can be used to augment drone flights acoustically to better differentiate airborne gestures or---in its simplest form---to attract attention.
In this way, a trajectory can convey more information, just as a shaky voice conveys additional auditory cues in a human dialogue.
A user study shows, that this method makes two similar-sounding trajectories more clearly distinguishable.

\subsection{Structure} 

The organization of this paper is as follows:
\Cref{sec:gestures} describes the drone platform used as well as the positive and negative feedback trajectories including the acoustic extension to better distinguish the negative feedback trajectory.
A detailed account of the user study is provided in \cref{sec:userstudy}.
\Cref{sec:discussion} discusses its results and limitations.
Finally, in \cref{sec:conclusion}, this paper is concluded and areas for future research are suggested.

%% file: 2_gestures.tex
\section{Airborne Gestures}\label{sec:gestures}

Two very basic human gestures to communicate positive and negative feedback in western civilizations are nodding and shaking one's head.
Transferring these movements to a drone results in similar sounding flights, despite their intent being opposite.
It can be challenging to identify them based on acoustic cues alone.
Especially in scenarios where a person cannot visually track a drone, audibly distinguishing its intent could enhance the interaction experience.

Augmenting the consequential sound of a drone in one of the trajectories promises improved differentiability.
One possibility to achieve this would be to adapt the controller used.
However, in order not to interfere with an established drone controller, the concrete implementation pursued in this approach is the alteration of the trajectories that are then fed to the controller.

This section describes the implementation of the basic positive and negative airborne feedback gestures that mimic human head nodding and head shaking.
It then addresses the augmentation by the proposed vocalics approach, as well as the drone platform used for the concrete implementation of the concept.

\subsection{Trajectories}

Quadrotors are underactuated systems, \ie, translational motion in the vertical plane is coupled to rotational motion about their roll and pitch axes.
For this reason, the human nod is mimicked by a simple up and down movement instead of actual nodding by a rotation about the drone's pitch axis, since this would introduce horizontal motion.
To communicate negative feedback, the drone repeatedly rotates left and right about its yaw axis to mimic human head shaking.

The trajectories are based on a simple harmonic motion:
\begin{equation*}
    h(t,f,p) = \sin(2\pi f \cdot t + p).
\end{equation*}
with time $t$, frequency $f$, and phase shift $p$.
For the positive feedback trajectory, scaling this function models a height offset
\begin{equation*}
    \dlt{z}(t) = 0.03 \cdot h(t, 1, \pi)
\end{equation*}
which is applied to the drone's current hovering height.
A phase shift of $p=\pi$ is applied to start the nodding motion downward first.
The head shaking motion used for the negative feedback trajectory has the same characteristics (thus includes the phase shift) and is scaled to
\begin{equation*}
    \dlt{\psi}(t) = \frac{\pi}{6} \cdot h(t, 1, \pi).
\end{equation*}

\begin{figure}
    \centering
    \begin{subfigure}[t]{1\textwidth}
        \includegraphics[width=\textwidth]{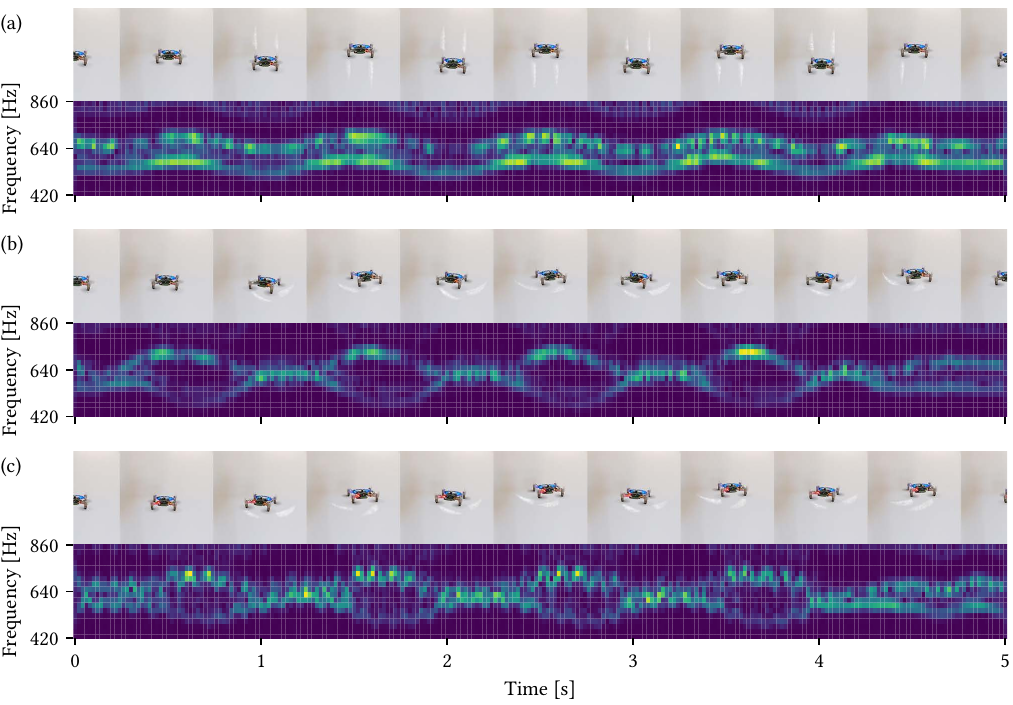}
        \phantomcaption
        \label{fig:stills_stft_positive}
    \end{subfigure}
    \begin{subfigure}[t]{0\textwidth}
         \phantomcaption
         \label{fig:stills_stft_negative_ord}
    \end{subfigure}
    \begin{subfigure}[t]{0\textwidth}
        \phantomcaption
        \label{fig:stills_stft_negative_voc}
   \end{subfigure}
    \captionsetup{subrefformat=parens}
    \caption{%
        Stills from trajectories communicating positive feedback \subref{fig:stills_stft_positive}, \emph{ordinary} negative feedback \subref{fig:stills_stft_negative_ord}, and \emph{vocalics} negative feedback \subref{fig:stills_stft_negative_voc} including relevant sections of the \acp*{stft} of their microphone recordings.
        The drone's movement is visually emphasized through illustrated traces.
        Since the rotors have two blades, the measured frequency corresponds to twice the rotor speed.
    }
    \label{fig:stills_stfts}
    \Description{%
        Image sequences of the drone flights that were performed in the user study and sections of the \acp*{stft} of the sound recordings.
        The positive feedback trajectory shows repeated up and down motions.
        The ordinary negative feedback trajectory and the vocalics feedback trajectory both show repeated yaw motions that cannot be distinguished.
        In the \acp*{stft} of the positive feedback trajectory and the ordinary negative feedback trajectory similar periodic waves are visible suggesting similar sounding flights.
        The \ac*{stft} of the vocalics negative feedback trajectory has the same periodicity as the ordinary negative feedback trajectory but with a superimposed and visible oscillating frequency.
    }
\end{figure}

Flying these trajectories using the drone described in \cref{sec:drone} results in audible sounds that are difficult to distinguish.
This similarity is visible in the \ac{stft} of a microphone recording of the performed flights and shown in \cref{fig:stills_stft_positive} and \cref{fig:stills_stft_negative_ord}.
The frequencies displayed correspond to the frequencies of the rotors.
If a quadrotor performs a translational motion along its vertical axis, thrust is equally applied to all four rotors, increasing their frequency (\cref{fig:stills_stft_positive}).
When a quadrotor spins about its vertical axis, more thrust is applied to one opposite pair of rotors, while the thrust to the other pair is reduced (\cref{fig:stills_stft_negative_ord}).

\begin{figure}
    \centering
    \begin{subfigure}[t]{\linewidth}
        \includegraphics[width=\linewidth]{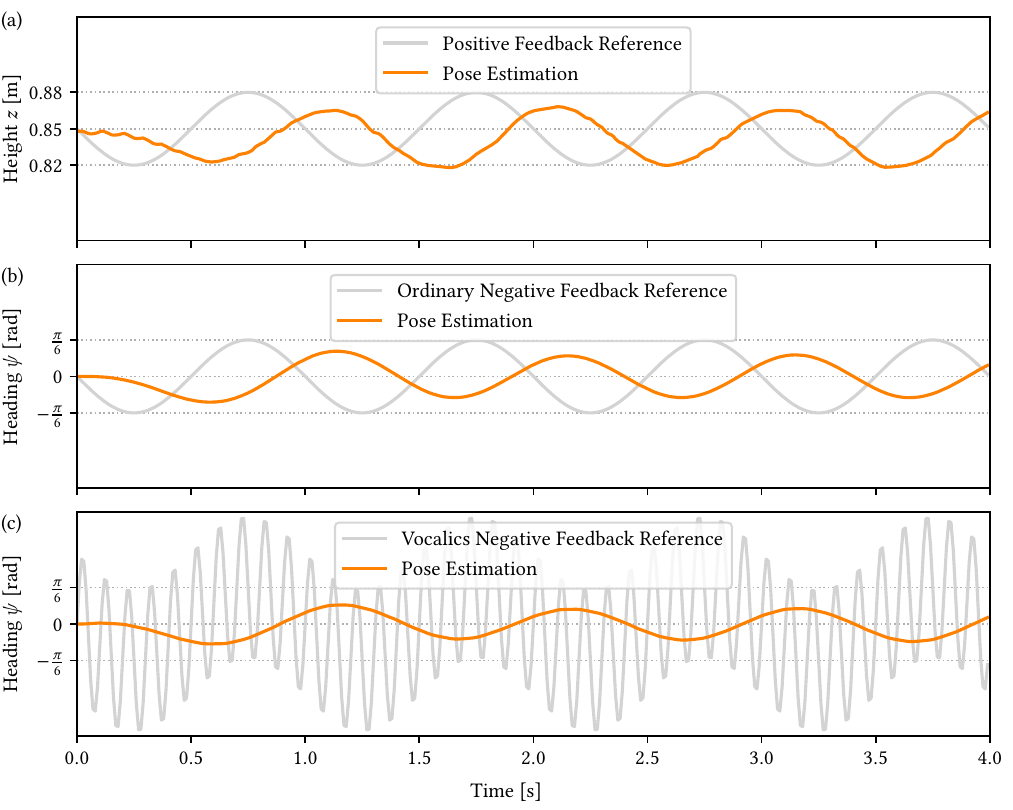}
        \phantomcaption
        \label{fig:coords_positive}
    \end{subfigure}
    \begin{subfigure}[t]{0\linewidth}
         \phantomcaption
         \label{fig:coords_negative_ord}
    \end{subfigure}
    \begin{subfigure}[t]{0\linewidth}
        \phantomcaption
        \label{fig:coords_negative_voc}
   \end{subfigure}
    \captionsetup{subrefformat=parens}
    \caption{%
        Relevant reference and tracked parameters of the three trajectories:
        The height $z$ for the positive feedback trajectory \subref{fig:coords_positive}, and the drone's yaw angle $\psi$ for both the ordinary negative feedback \subref{fig:coords_negative_ord} and the vocalics negative feedback \subref{fig:coords_negative_voc} trajectories.
        The background represents the corresponding reference parameter, while the parameter tracked by the pose estimation system is shown in orange.
    }
    \Description{%
        Plots of the positive feedback trajectory, the ordinary negative feedback trajectory, and the vocalics negative feedback trajectory, all of them with overlaid tracking data from the flights that where conducted for the user study.
    }
    \label{fig:coords}
\end{figure}

Due to the similarity in applied thrust, both flights sound very similar.
In order to be able to better distinguish the negative feedback trajectory from the positive feedback trajectory,
the negative feedback trajectory was superimposed by an additional \SI{10}{\hertz} harmonic motion
\begin{equation*}
    \dlt{\psi}_\text{voc}(t) = \dlt{\psi}(t) + \frac{\pi}{3} \cdot h(t,10,0)
\end{equation*}
to obtain the \emph{vocalics} negative feedback trajectory.

The \ac{stft} of the sound recorded during its flight is shown in \cref{fig:stills_stft_negative_voc}.
The relevant coordinates of both the \emph{ordinary} negative feedback trajectory and the \emph{vocalics} negative feedback trajectory are shown in \cref{fig:coords_negative_ord} and \cref{fig:coords_negative_voc}, respectively.
The displayed heading angle obtained by the pose estimation system during flights does not reflect the oscillation applied to the reference trajectory.
This emphasizes that the trajectories are visually indistinguishable from each other, despite superimposing a higher frequency signal that is clearly audible.

\subsection{Drone Platform}\label{sec:drone}

\begin{figure}
    \centering
    \includegraphics{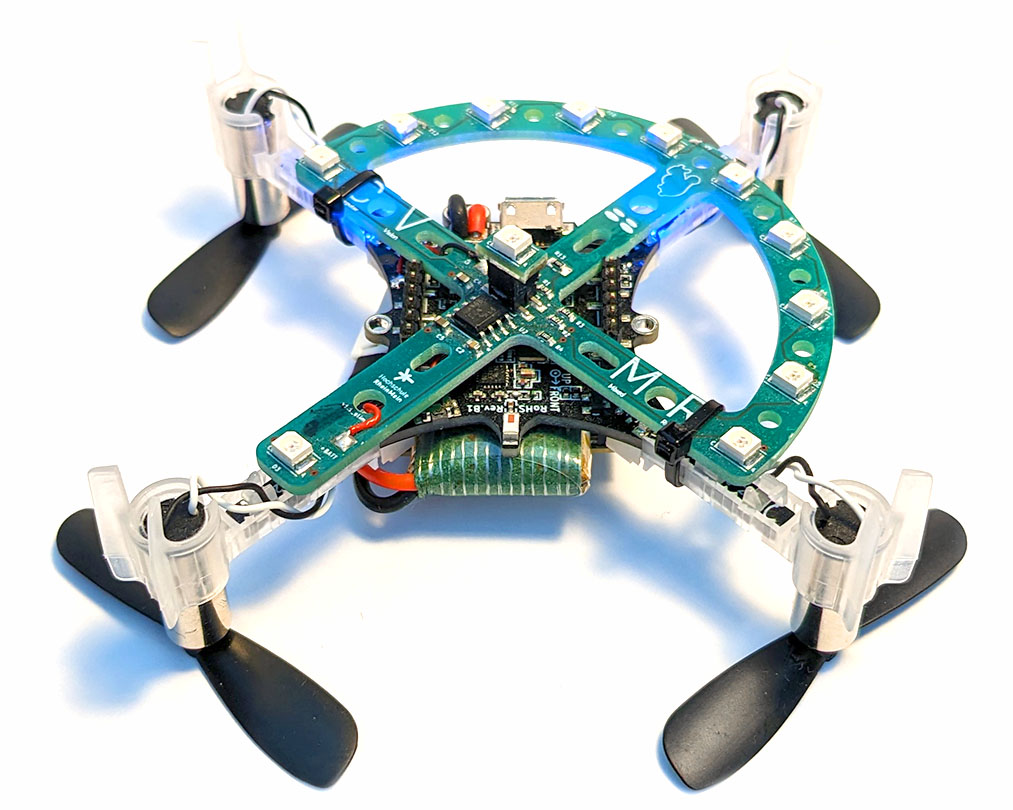}
    \caption{%
    The Bitcraze Crazyflie 2.1 development drone platform used for this user study, with its rotors mounted upside down to avoid obscuring the LEDs of the tracking marker.
    }
    \Description{A small drone platform with an attached tracking marker.}
    \label{fig:crazyflie}
\end{figure}

For this experiment, the slightly modified Bitcraze Crazyflie 2.1 platform shown in \cref{fig:crazyflie} is used.
It is controlled from a regular laptop running an implementation of the widespread control algorithm described by \citet{Mellinger2011} using a Crazyradio PA \SI{2.4}{\giga\hertz} USB dongle.
The quadrotor is turned upside down to be extended by an active infrared LED tracking marker.
This marker is tracked by a monocular pose estimation system~\cite{tjaden2019} and with the specific implementation provides poses with six \acl{dof} at a rate of \SI{100}{\hertz}.
In the specific setup for this paper, a single Ximea MQ013MG-ON USB3 high-speed vision camera with a resolution of \qtyproduct[product-units=single]{1280x1024}{\pixel} is used in combination with a Fujinon DF6HA-1B 1:1.2/\SI[parse-numbers=false]{6}{\milli\meter} lense and a MidOpt FIL BN850/27 bandpass filter matching the LEDs' range of wavelengths.
To reduce noise and jitter, a first-order filter with time constant $\tau=\SI{0.05}{\second}$ is used to smooth poses over time.
This is part of an affordable testbed developed over recent years, enabling high-speed drone maneuvers with high accuracy.
Nevertheless, for implementing similar studies that use simple trajectories, such as the ones used in this specific experiment, there are ready-to-use commercial pose estimation solutions available for the Crazyflie.

%% file: 3_userstudy.tex
\section{User Study}\label{sec:userstudy}

In order to evaluate whether the vocalic extension of gestures can add value in \ac{hdi} by helping to audibly differentiate similar sounding trajectories, the flights of the positive feedback trajectory, the \emph{ordinary} negative feedback trajectory, and the \emph{vocalics} negative feedback trajectory were recorded.
Video and sound recordings were presented separately within an online survey to two groups of voluntary participants.
The study was implemented using the open-source platform oTree~\cite{otree2016} and is described in the following.

\subsection{Multimedia Recordings and Synchronization}

The video was recorded using a Google Pixel 5 with default camera settings.
The audio was captured with a Shure SM57 instrument microphone equipped with a noise protection filter and a Focusrite Scarlett 2i2 studio-quality interface.
Audio and video of the individual airborne gestures were synchronized with a signal tone that was played mid-air just before the start of the gesture.
For later evaluation, the software controlling the drone logged important variables at \SI{100}{\hertz}, such as control input, position, attitude, battery voltage, and control method.
The latter changes when switching from hover control to trajectory control used for the airborne gestures.
This switch is used to synchronize the log data with the audio recording and the signal tone mentioned before.

\subsection{Hypotheses}

The alternative hypothesis is that \emph{the group whose negative feedback gestures were acoustically augmented by vocalics can better link the sounds to the videos}, implying higher audible information content in the drone's \emph{vocalics} negative feedback trajectory, as it is visually almost indistinguishable from the \emph{ordinary} negative feedback trajectory.
To reject the null hypothesis that \emph{the additional information will not increase the chance of linking the sound correctly to the gestures}, the common significance level of \SI{5}{\percent} is chosen.

\subsection{Study Design}\label{sec:design}

The participants were shown a brief written introduction to the user study and a consent form.
It was stated, that in this study, they would be shown two short videos, each of which shows a drone performing a specific movement.
Afterwards, two audio files would be played for them.
Furthermore it was stated, that it would be a prerequisite for this study that their device has an audio output and that they should make sure, that the media sound was enabled.
After agreeing to the consent form, age, gender, and occupation were asked before the actual experiment started.

The user study was divided into two groups: The \emph{ordinary} trajectory group and the \emph{vocalics} trajectory group.
Both groups would see two videos and then listen to two audio files of the same drone flights: the positive feedback gesture displayed in \cref{fig:stills_stft_positive} and either the \emph{ordinary} negative feedback trajectory shown in \cref{fig:stills_stft_negative_ord} or the \emph{vocalics} negative feedback trajectory displayed in \cref{fig:stills_stft_negative_voc}.
The videos presented to participants in the first part of the study (from which the audio for the second part of the study was extracted) are available on YouTube\footnote{\url{https://www.youtube.com/playlist?list=PLPhCA5Y9lesmAPD5TRXISybvvMxNVeOjA}}.
To ensure impartiality, the participants of both groups were again divided into four subgroups of 24 participants each so that the media could be presented in the pseudorandom sequences shown in \cref{tab:media_order}.

\begin{table}[!hbp]
    \caption{Media orders of subgroups where media index 1 is the positive feedback trajectory and media index 2 is the negative feedback trajectory.}
    \label{tab:media_order}
    \centering
    \begin{tabular}{@{}ccccc@{}}
        \toprule
        Subgroup & \multicolumn{4}{c}{Survey Media Order} \\
        & 1 & 2 & 3 & 4 \\
        \cmidrule(lr){2-5}
        & Video & Video & Sound & Sound \\
        \midrule
        1 & 1 & 2 & 1 & 2\\
        2 & 1 & 2 & 2 & 1\\
        3 & 2 & 1 & 1 & 2\\
        4 & 2 & 1 & 2 & 1\\
        \bottomrule
    \end{tabular}
\end{table}

\noindent
The participants could watch the videos multiple times or listen to the sounds multiple times, but they could not navigate back to the previous media item.
The buttons to move to the next page were disabled until the media playback finished.
This ensured that the participants played the media at all, and all the way to the end.
The data of the participants who have not reached the last page of the study have been deleted.
After each of the two audio files, the participants should link the sound they just heard to one of the two previously shown videos.
They had the choice between \emph{Video 1}, \emph{Video 2}, and \emph{Not sure}.
Both sounds had to be correctly assigned to the corresponding videos to be considered \emph{correctly linked}.

\subsection{Participants}

The user study ($N=192$) was conducted with 99 males (\SI{51.6}{\percent}), 90 females (\SI{46.9}{\percent}), three diverse or undisclosed (\SI{1.5}{\percent}) participants aged 19 to 69 years (mean $\bar{x}=36.0$, standard deviation $\sigma=7.8$).
They were asked to voluntarily take part in the study via social media.

\subsection{Results}

Out of the 96 participants in the \emph{ordinary} group, 44 (\SI{45.83}{\percent}) could link both sounds correctly to the gestures whereas for the \emph{vocalics} group, 64 of the 96 participants (\SI{66.67}{\percent}) were able to do so.
The collected data are tabulated as the $2\times2$ contingency \cref{tab:contingency_table}.

\begin{table}[!htb]
    \caption{Contingency table presenting the results of the conducted user study.}
    \label{tab:contingency_table}
    \centering
    \begin{tabular}{@{}cS[table-format=2.0,table-column-width=14mm]S[table-format=2.0,table-column-width=14mm]S[table-format=3.0,table-column-width=14mm]@{}}
        \toprule
        Correctly & {Ordinary} & {Vocalics} & {Combined}\\
        linked & {(Group A)} & {(Group B)} & {response}\\
        \midrule
        Yes & 44 & 64 & 108\\
        No & 52 & 32 &  84\\
        \midrule
        Total & 96 & 96 & 192\\
        \bottomrule
    \end{tabular}
\end{table}

\noindent
To evaluate the contingency table, \texttt{SciPy}'s~\cite{SciPy2020} implementation of the Barnard's exact test~\cite{barnard1947} is used.
Under the null hypothesis that the additional information introduced by the vocalics will not increase the chance of correct linkage to the gestures, the likelihood of obtaining test outcomes at least as extreme as the observed data is approximately \SI{0.19}{\percent}.
As the $p$-value is below the chosen significance level of $0.05$, there is sufficient evidence to reject the null hypothesis in favor of the alternative hypothesis.

%% file: 4_discussion.tex
\section{Discussion}\label{sec:discussion}

\subsection{Results}

The results of the experiment suggest that the superimposition of a frequency to the flight sound in the \emph{vocalics} group had a positive impact on participants' ability to connect the flight sounds to the aerial drone gestures.
The results indicate that sonically augmenting the trajectory in the \emph{vocalics} group made it easier for participants to associate and differentiate between the flight sounds and gestures, resulting in a higher success rate in linking them compared to the \emph{ordinary} group.
This underlines that the idea of extending the consequential sound of drones is capable of conveying information and thus opens up a previously unexplored channel of nonverbal communication in \ac{hdi}.

\subsection{Methodology}

Online studies facilitate reaching a larger number of individuals across a broad demographic spectrum.
More importantly, impacts typically associated with drone proxemics, such as increased mental stress, arise from the physical presence of the drone.
Through an online study, it was ensured that the results were free from these influences.
Participants were able to fully focus on the visual and acoustic content of the study in an environment of their own choice.

The online study was conducted independently by the participants without supervision.
Control over the playback of multimedia content was thus limited.
As described in \cref{sec:design}, measures were taken to ensure that the videos and sounds were played to completion.
However, browsers do not have access to system settings such as volume.
It is relied upon that the users, who have all voluntarily participated, have indeed turned the sound on as requested on the first slide of the survey.
There was also no control, if the participants used their device's speaker, headphones or external speakers.
It is presumed that participants are familiar with their own devices and consume media through them, as they were acquired through social media.
If they did not change their surroundings between individual aerial gestures, there are no concerns in this regard for this study as they were meant to compare sounds with each other; it is only important that they would hear them through the same device and in the same surrounding.
There were two participants (z score > 2) who took more than the average \SI{2.7}{\minute} for the four slides containing media (namely \SI{44}{\minute} and \SI{264}{\minute}), for whom it could be doubted whether they had consumed all media in the same environment.
Excluding these two participants from the statistical analysis further decreases the $p$-value to $p=0.0017$.

\subsection{Drone Vocalics Limitations}

The augmentation of drone rotor sounds as part of \ac{hri} is limited to ranges within earshot.
Loud environments, where ambient noise drowns out the drone sound are also unsuitable for drone vocalics as an isolated communication channel.
On larger drone platforms with slower rotor speeds, it remains to be investigated how the proposed method can be implemented without affecting flight characteristics.
The impact of superimposing the trajectory with a higher frequency signal on the quadrotor battery also remains uncertain and requires further evaluation.

\subsection{Technical Limitations}

The yaw movement of the drone induced a noticeable change in altitude in both the \emph{ordinary} and \emph{vocalics} negative feedback trajectories (\cref{fig:stills_stft_negative_ord,fig:stills_stft_negative_voc}).
Rotation about the yaw axis is achieved by torque imbalance.
In a quadrotor, opposing rotors spin in the same direction: one pair rotates clockwise, while the other pair rotates counterclockwise.
To initiate yaw motion, the speed of rotors spinning in one direction is increased, while the speed of rotors spinning in the opposite direction is decreased.
This results in a net torque, causing angular acceleration in the desired direction.
As rotors spin up faster than they spin down, the sum of the forces generated by the individual rotors during the yaw movement can momentarily exceed the thrust necessary to maintain the current altitude.
Additionally, the flight controller can decrease rotor speeds only to a minimum idling speed contributing to this effect.

There is a latency of the flown trajectories to their reference trajectories visible in \cref{fig:coords}.
This has no effects on the study or its results, but the latency should be further reduced by optimizing the smoothing of the pose estimation system and controller parameters.

The total rotation angle of the yaw motion was set to \SI{60}{\degree} in the reference negative feedback trajectory.
As the controller used does not take the mathematical model parameters of the specific drone and the feasibility of trajectories into account, the \emph{ordinary} negative feedback trajectory only reaches a full head shaking range of \SI{42}{\degree}.
The overlaid signal in the \emph{vocalics} negative feedback trajectory further reduces this range to \SI{32}{\degree} and could be compensated by targeting a larger angle.

%% file: 5_conclusion.tex
\section{Conclusion \& Future Work}\label{sec:conclusion}

In this paper, a user study was conducted to evaluate the potential of adding audible information to drone trajectories.
Positive and negative feedback trajectories mimicking human gestures were generated and tracked by a quadrotor resulting in visually clearly differentiable flights.
However, in situations where keeping the drone in sight is not possible or for visually impaired people, distinguishing airborne gestures acoustically becomes challenging.
Superimposing a higher frequency signal to the negative feedback trajectory did not affect the visual impression of the flight, but could add enough acoustic information to be better differentiable from the positive feedback trajectory.
Next to the results and methodology of the user study, the technical limitations have been discussed.
Enhancements of the used infrastructure include optimizing pose estimation and controller parameters to further reduce lag and stabilize the drone's altitude during yaw movements.

In conclusion, this study represents an initial step toward integrating vocalics into \ac{hdi} by adding auditory cues utilizing onboard means.
Building on this, new ways to further improve \ac{hdi} can be explored, \eg, by developing new interface designs that include both visual and auditory cues.
Especially for individuals with visual impairments, this could make drone technology more accessible.
\ac{hri} thrives by incorporating as many nonverbal channels as possible.
Providing the community with the drone vocalics concept offers a tool that can be used to further enhance the interaction experience.
In future scenarios, where drones and their sounds may have become more commonplace, vocalics could be an additional channel for attracting attention alongside the currently available possibilities, or to communicate with pedestrians.
The proposed method can be expanded by superimposing different frequencies, encoding sequences of varying signal durations, such as Morse code, or mimicking bio-inspired sounds like the pulsed vibration used by honey bees for communication.
This study served as a proof of concept; future studies will focus on exploring a wider variety of tones, communicating more meaningful messages and emotions, and qualitatively assessing them.